\documentclass[letterpaper, 10pt, conference]{IEEEtran}
\usepackage[backend=biber, style=ieee]{biblatex}
\usepackage{graphicx}
\usepackage{amsmath,amssymb,amsfonts}
\usepackage{booktabs}
\usepackage{multirow}
\usepackage{xcolor}
\usepackage[colorlinks=true, allcolors=blue]{hyperref}

\newcommand{\method}{EIPNet}

\title{\LARGE\bf \method{}: Leveraging Inertial Measurements for
Event-based 6-DoF Camera Relocalization}

\author{
  \IEEEauthorblockN{Ahmed Tabia}
 \IEEEauthorblockA{\begin{tabular}[t]{c}ESIEE Paris, Universit\'e Gustave Eiffel, France\\ \texttt{ahmed.tabia@esiee.fr}\end{tabular}}
  
}

\begin{document}
\maketitle
\thispagestyle{empty}
\pagestyle{empty}
\begin{center}\footnotesize\itshape
This work has been submitted to the IEEE for possible publication. Copyright may be transferred without notice, after which this version may no longer be accessible.
\end{center}

\begin{abstract}
Event cameras are attractive sensors for 6-DoF camera relocalization
thanks to their high temporal resolution, high dynamic range, and
robustness to motion blur. Most event platforms, including the
DAVIS240C used in standard benchmarks, embed a synchronized inertial
measurement unit (IMU); yet, to date, every learning-based
relocalization method discards it and regresses the pose from the
event stream alone. In this letter, we revisit this design choice and
quantify how much this freely available inertial modality can improve
relocalization. We propose \method{}, a lightweight event-inertial
network that encodes events as a raw point cloud and fuses them with
an inertial branch. On the DAVIS RPG benchmark, fusing the IMU reduces
the median position and orientation errors by $58\%$ and $43\%$
respectively, allowing a $0.38$M-parameter model to match or surpass
the state-of-the-art PEPNet while using half of its parameters and
running in real time on a CPU. An ablation study further reveals that
a simple feature concatenation captures the entire gain, a more
elaborate cross-attention fusion bringing no measurable improvement.
\end{abstract}

\section{INTRODUCTION}
\label{sec:intro}

Event cameras are bio-inspired vision sensors that asynchronously
report per-pixel brightness changes with microsecond latency, rather
than capturing synchronous intensity frames at a fixed
rate~\cite{mueggler2017dataset}. This operating principle grants them
a high dynamic range, very low latency, and robustness to motion blur,
making them particularly well suited to perception under high-speed
motion and challenging illumination. Combined with their low power
consumption, these properties have made event cameras an appealing
choice for resource-constrained robotic and augmented-reality
platforms.

Among the tasks they enable, 6-DoF camera relocalization recovering
the absolute pose of the camera within a known scene is a fundamental building block for navigation, manipulation, and augmented
reality. The asynchronous event stream implicitly encodes fine-grained
motion and depth cues, which makes event cameras attractive for this
task. However, exploiting these cues effectively is non-trivial: a
single event carries very little information, and the way the event
stream is represented strongly conditions the accuracy of the
downstream pose estimate.

A large body of learning-based methods addresses event-based
relocalization using \emph{only} the event stream.
PoseNet~\cite{kendall2015posenet} and its Bayesian
variant~\cite{kendall2016bayesian} pioneered CNN-based absolute pose
regression and were transferred to event data by
SP-LSTM~\cite{nguyen2019splstm}, which forms event images and learns
spatial dependencies with stacked spatial LSTMs. More recent methods
further improve accuracy, including the CNN-LSTM of
Tabia et al.~\cite{tabia2022deep} and the point-based
PEPNet~\cite{pepnet2024}, the latter processing events as a raw point
cloud to preserve their temporal resolution. Despite their
architectural differences, all these methods share one assumption:
the pose must be inferred from events alone. SP-LSTM even explicitly
presents the absence of inertial data as a design feature.

This assumption overlooks a resource that is physically present on
virtually every event-camera platform. The DAVIS240C used to record
the standard benchmark sequences embeds a \emph{synchronized inertial
measurement unit (IMU)}, which directly measures angular velocity and
linear acceleration. In particular, the accelerometer senses the
gravity vector, providing an absolute cue on the camera's roll and
pitch that no amount of visual processing can recover from events
alone. Inertial data is routinely exploited in visual-inertial
odometry, yet it has been systematically discarded in learning-based
event relocalization. This raises a simple question that, to our
knowledge, has not been answered on this benchmark: \emph{how much can
this freely available inertial modality improve event-based
relocalization, and at what computational cost?}

We answer this question with \method{} (Event-Inertial Point Network),
a lightweight architecture that encodes events as a raw point cloud
and fuses them with a compact inertial branch. We deliberately keep
the design minimal, in line with the energy-efficiency rationale of
event cameras, and show that this simplicity is sufficient to reach
state-of-the-art accuracy. Our main contributions are:

\begin{itemize}
  \item We propose \method{}, to our knowledge the first
  learning-based method to exploit inertial measurements for absolute
  6-DoF event-camera relocalization, using a point-cloud event encoder
  and an inertial branch fused by simple concatenation.

  \item We provide the first quantitative measurement of the inertial
  contribution on the DAVIS RPG benchmark. Under identical settings,
  fusing the IMU reduces the median position and orientation errors by
  $58\%$ and $43\%$ respectively on the \texttt{shapes\_rotation}
  sequence, and lets a $0.38$M-parameter model match or surpass the
  state-of-the-art PEPNet while using half of its parameters.

  \item Through an ablation study, we show that a simple concatenation
  of the two modalities captures the entire gain, whereas a more
  elaborate cross-attention fusion brings no measurable improvement
  despite doubling the parameter count---a finding that justifies our
  lightweight design.
\end{itemize}

\section{RELATED WORK}
\label{sec:related}

\subsection{Learning-based Event Relocalization}
Camera pose relocalization with deep networks was pioneered by
PoseNet~\cite{kendall2015posenet}, which regresses an absolute 6-DoF
pose from a single RGB image using a convolutional backbone. Its
Bayesian extension~\cite{kendall2016bayesian} adds uncertainty
estimation, and a geometric reprojection loss~\cite{kendall2017geo}
later improved rotation accuracy. These frame-based regressors were
transferred to event cameras by SP-LSTM~\cite{nguyen2019splstm}, which
accumulates events into event images and captures spatial dependencies
with stacked spatial LSTMs, substantially outperforming direct PoseNet
baselines on event data. Subsequent work pushed accuracy further: the CNN-LSTM of
Tabia et al.~\cite{tabia2022deep} combines convolutional features
with a recurrent module, while PEPNet~\cite{pepnet2024} discards the
image abstraction altogether and processes the raw event cloud with a
hierarchical point-based encoder and an attentive bidirectional LSTM,
reaching state-of-the-art accuracy with a remarkably small parameter
budget. All of these methods, however, regress the pose from the event
stream alone and leave the on-board inertial sensor unused.

\subsection{Event Representations}
The way the asynchronous event stream is encoded strongly conditions
the accuracy of any downstream task. The most common choice is to
accumulate events over a short time interval into a 2-D \emph{event
image}~\cite{nguyen2019splstm}, which enables standard convolutional
backbones but collapses the precise timestamp of each event.
\emph{Voxel grids} and related representations mitigate this loss by
discretizing events into a small number of temporal
bins~\cite{gehrig2019end}, trading temporal resolution for a
regular tensor structure. More recently, point-based representations
treat the event stream as a sparse point cloud
$(x, y, t, \rho)$~\cite{pepnet2024}, preserving the full microsecond
resolution and avoiding any spatial or temporal quantization. This
representation is naturally aligned with the sparse, asynchronous
nature of event data and admits lightweight permutation-invariant
encoders in the spirit of PointNet~\cite{qi2017pointnet}. We adopt
this point-based representation for its temporal fidelity and its
efficiency.

\subsection{Visual-Inertial Fusion}
Fusing visual and inertial measurements is a long-standing strategy in
odometry and SLAM. Filtering-based methods such as the
MSCKF~\cite{mourikis2007msckf} and tightly-coupled optimization
frameworks such as VINS-Mono~\cite{qin2018vinsmono} combine camera and
IMU data to achieve robust state estimation under fast motion and
textureless scenes. In the event-camera domain, Ultimate
SLAM~\cite{vidal2018ultimate} fuses events, standard frames, and
inertial data to track the 6-DoF pose in high-speed and
high-dynamic-range scenarios, and learned inertial odometry has also
been explored for IMU-only ego-motion estimation. These approaches,
however, target \emph{incremental} ego-motion or tracking rather than
\emph{absolute} relocalization in a known scene, and none combine a
point-cloud event representation with inertial data. To our knowledge,
\method{} is the first learning-based method to exploit the IMU for
absolute 6-DoF event-camera relocalization, closing a gap that prior
point-based work~\cite{pepnet2024} left open.

\section{METHOD}
\label{sec:method}

\begin{figure*}[t]
  \centering
  \includegraphics[width=\textwidth]{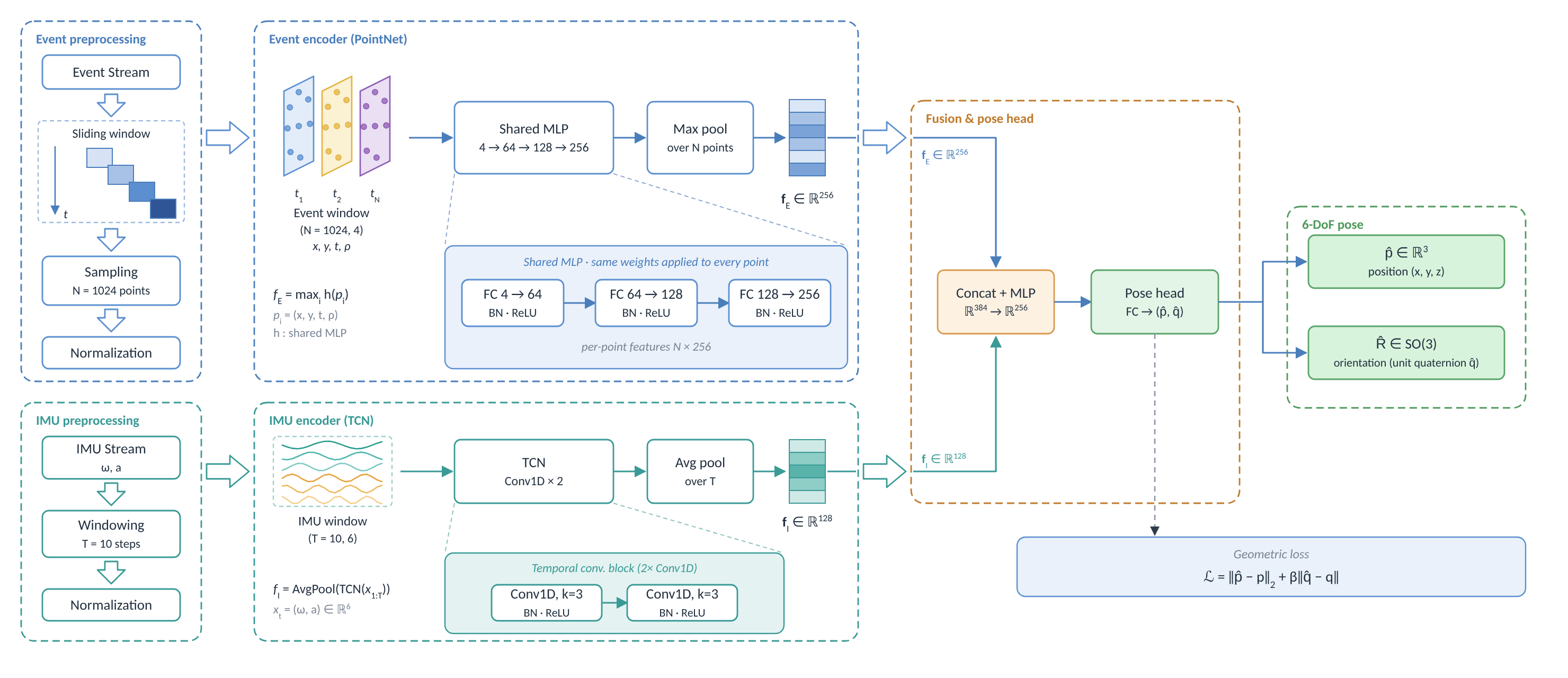}
  \caption{Overview of \method{}. A short window of events is encoded
  as a 4-D point cloud by a PointNet-style encoder, while the
  synchronized IMU window is encoded by a temporal convolutional
  network. The two modality features are fused by concatenation and
  regressed to an absolute 6-DoF pose through a shared head.}
  \label{fig:arch}
\end{figure*}

\subsection{Problem Formulation}
We cast event-based relocalization as an absolute pose regression
problem. Following the protocol of~\cite{nguyen2019splstm}, one sample
is defined by the set of events occurring between two consecutive
motion-capture timestamps. Since the ground-truth system operates at
$200$~Hz, each sample corresponds to a temporal window of
approximately $5$~ms. For a window ending at time $t$, the network
receives the events and the inertial measurements falling in that
window, and regresses the absolute pose
$\mathbf{y}=[\mathbf{p},\mathbf{q}]$ of the camera, where
$\mathbf{p}\in\mathbb{R}^3$ is the position and
$\mathbf{q}\in\mathbb{R}^4$ the unit quaternion encoding orientation.
The target pose is taken as the motion-capture measurement at the end
of the window, which is available at sub-millimeter precision and
requires no interpolation.

\subsection{Event Representation}
Rather than accumulating events into a 2-D image, we preserve their
temporal structure by treating the stream as a point cloud. Each event
$e=(x,y,t,\rho)$, with pixel coordinates $(x,y)$, timestamp $t$ and
polarity $\rho\in\{0,1\}$, is mapped to a normalized 4-D point: the
spatial coordinates are divided by the sensor resolution, the
timestamp is normalized to $[0,1]$ within the window, and the polarity
is kept as a binary channel. From each window we sample $N=1024$ events
(with replacement when fewer are available) to form a fixed-size point
cloud $\mathcal{P}=\{\mathbf{p}_i\}_{i=1}^{N}\in\mathbb{R}^{N\times4}$.
This representation retains the full temporal resolution of the stream
and avoids any spatial or temporal quantization.

\subsection{Event Encoder}
We encode the point cloud with a lightweight PointNet-style
network~\cite{qi2017pointnet}. A shared multi-layer perceptron (MLP),
applied identically to every point, lifts each 4-D point to a
high-dimensional feature through three fully connected layers of
dimensions $4\!\to\!64\!\to\!128\!\to\!256$, each followed by batch
normalization and a ReLU activation. A symmetric max-pooling across
the $N$ points then aggregates the per-point features into a single
order-invariant descriptor $\mathbf{f}_E\in\mathbb{R}^{256}$:
\begin{equation}
  \mathbf{f}_E = \max_{i=1,\dots,N} h(\mathbf{p}_i),
\end{equation}
where $h$ denotes the shared MLP. The max-pooling makes the encoder
invariant to the ordering of the sampled events, which is essential
since the point cloud carries no canonical order.

\subsection{Inertial Encoder}
The inertial branch processes the IMU window
$\mathcal{I}\in\mathbb{R}^{T\times6}$, where $T=10$ resampled steps
each contain the three-axis gyroscope and three-axis accelerometer
readings. We use a temporal convolutional encoder rather than a
recurrent one: two 1-D convolutional layers (kernel size $3$, channel
widths $6\!\to\!64\!\to\!64$), each followed by ReLU, capture the local
temporal dynamics, and an adaptive average-pooling over the time axis
produces a fixed descriptor that a small MLP projects to
$\mathbf{f}_I\in\mathbb{R}^{128}$. Compared to an LSTM or GRU, this
design is fully parallel over the time axis, has fewer parameters, and
avoids recurrent dependencies that are unnecessary for such short
windows, while remaining hardware-friendly.

\subsection{Fusion}
The two modality descriptors are combined by concatenation followed by
a projection MLP:
\begin{equation}
  \mathbf{f} = \mathrm{MLP}\big([\,\mathbf{f}_E \,;\, \mathbf{f}_I\,]\big)
  \in \mathbb{R}^{256},
\end{equation}
mapping the concatenated $\mathbb{R}^{384}$ vector to a fused
$\mathbb{R}^{256}$ representation, with layer normalization, a ReLU and
dropout. We deliberately adopt this minimal fusion; we also
experimented with a cross-attention mechanism between the two
modalities, but found that it brings no measurable improvement while
doubling the parameter count (Table~\ref{tab:ablation}).

\subsection{Pose Head and Loss}
The fused feature is decoded by two parallel heads. A position head
regresses $\hat{\mathbf{p}}\in\mathbb{R}^3$ directly. An orientation
head regresses a continuous 6-D rotation
representation~\cite{zhou2019continuity}, which is converted to a valid
rotation matrix $\hat{\mathbf{R}}\in SO(3)$ by Gram--Schmidt
orthogonalization and then to a quaternion $\hat{\mathbf{q}}$ for the
loss. We avoid regressing quaternions directly, as the discontinuity
of the quaternion representation of $SO(3)$ hampers gradient-based
optimization. The network is trained with a geometric loss that
balances translation and rotation through a single learned
weight~\cite{kendall2017geo}:
\begin{equation}
  \mathcal{L} = \|\hat{\mathbf{p}}-\mathbf{p}\|_2
  + \beta\,\min\!\big(\|\hat{\mathbf{q}}-\mathbf{q}\|_2,
  \|\hat{\mathbf{q}}+\mathbf{q}\|_2\big),
\end{equation}
where the $\min$ resolves the sign ambiguity of unit quaternions
($\mathbf{q}$ and $-\mathbf{q}$ denote the same rotation) and
$\beta=e^{s}$, with $s$ a learned scalar, removes the need to tune the
relative weighting of position and orientation by hand.

\section{EXPERIMENTS}
\label{sec:exp}

\subsection{Dataset and Protocol}
We evaluate on the DAVIS RPG event-camera
dataset~\cite{mueggler2017dataset}, using six sequences and the
\emph{random split} protocol of~\cite{nguyen2019splstm}: $70\%$ of the
windows are used for training and $30\%$ for testing, with a fixed
random seed. For computational reasons, we uniformly subsample
$6{,}000$ windows per sequence ($3{,}000$ for the two densest
sequences, \texttt{hdr\_poster} and \texttt{poster\_translation}); the
$70/30$ split and the $5$~ms window definition are otherwise unchanged.

\subsection{Implementation Details}
\method{} is implemented in PyTorch and trained with AdamW~\cite{loshchilov2019adamw} ($\text{lr}=10^{-4}$, weight decay
$10^{-4}$) and a cosine-annealing schedule. Each model is trained until
convergence ($40$--$45$ epochs depending on the sequence) on a single
CPU. We report, for each sequence, the median position error (in
meters) and median orientation error (in degrees) on the test split,
following the evaluation protocol of~\cite{nguyen2019splstm}.

\begin{table*}[t]
\caption{Median error on the DAVIS RPG benchmark, random split
protocol. Position in meters, orientation in degrees. Best result in
\textbf{bold}. Input column: E = events only, E+I = events + IMU.}
\label{tab:random}
\centering
\small
\begin{tabular}{lcccccc}
\toprule
Sequence & PoseNet & Bayesian PN & SP-LSTM &
CNN-LSTM~\cite{tabia2022deep} & PEPNet & \textbf{\method{} (Ours)} \\
Input & E & E & E & E & E & \textbf{E+I} \\
Params & 12.43M & 22.35M & 135.25M & 12.63M & 0.774M & \textbf{0.379M} \\
\midrule
shapes\_rotation
 & 0.109, 7.388 & 0.142, 9.557 & 0.025, 2.256 & 0.012, 1.652
 & 0.005, 1.372 & \textbf{0.005, 1.009} \\
box\_translation
 & 0.193, 6.977 & 0.190, 6.636 & 0.036, 2.195 & 0.013, 0.873
 & 0.017, 0.845 & \textbf{0.015, 0.830} \\
shapes\_translation
 & 0.238, 6.001 & 0.264, 6.235 & 0.035, 2.117 & 0.020, 1.471
 & \textbf{0.011, 0.582} & 0.011, 0.727 \\
dynamic\_6dof
 & 0.297, 9.332 & 0.296, 8.963 & 0.031, 2.047 & 0.016, 1.662
 & 0.015, 1.045 & \textbf{0.016, 0.912} \\
hdr\_poster
 & 0.282, 8.513 & 0.290, 8.710 & 0.051, 3.354 & 0.033, 2.421
 & \textbf{0.016, 0.991} & 0.016, 1.125 \\
poster\_translation
 & 0.266, 6.516 & 0.264, 5.459 & 0.036, 2.074 & 0.020, 1.468
 & 0.012, 0.588 & \textbf{0.010, 0.432} \\
\midrule
Average
 & 0.231, 7.455 & 0.241, 7.593 & 0.036, 2.341 & 0.019, 1.591
 & 0.013, 0.904 & \textbf{0.012, 0.839} \\
\bottomrule
\end{tabular}
\end{table*}

\subsection{Comparison with the State of the Art}
Table~\ref{tab:random} reports the median position and orientation
errors on the six sequences of the random-split benchmark, together
with the parameter count of each method. \method{} is the only
approach that exploits the inertial modality, and it achieves the best
or second-best accuracy on every sequence while being by far the most
compact model. On \texttt{shapes\_rotation},
\texttt{box\_translation}, \texttt{dynamic\_6dof} and
\texttt{poster\_translation}, \method{} improves over the
state-of-the-art PEPNet in orientation and matches it in position.
Averaged over all sequences, \method{} reaches $0.012$~m and
$0.839^\circ$, slightly ahead of PEPNet ($0.013$~m, $0.904^\circ$)
while using only $0.379$M parameters, roughly half of PEPNet's
$0.774$M and two orders of magnitude fewer than SP-LSTM. We note that
PEPNet retains an edge on \texttt{shapes\_translation} and
\texttt{hdr\_poster}; the latter is the most cluttered scene of the
benchmark, where the event images are noisiest and the pose is hardest
to recover for all methods. Overall, these results show that a
deliberately simple event-inertial model is competitive with a
carefully engineered point-based architecture, and that the inertial
modality compensates for the absence of a hierarchical spatial encoder.

\begin{table}[t]
\caption{Ablation on the shapes\_rotation sequence (identical settings).
The inertial modality drives the entire gain; cross-attention adds
nothing.}
\label{tab:ablation}
\centering
\small
\begin{tabular}{lcccc}
\toprule
Variant & Input & Fusion & Params & Pos.~/~Ori. \\
\midrule
A & E   & ---            & 0.307M & 0.0112 / 1.780 \\
\textbf{B} & \textbf{E+I} & \textbf{concat} & \textbf{0.379M}
  & \textbf{0.0047 / 1.009} \\
C & E+I & cross-attn     & 0.807M & 0.0048 / 1.009 \\
\bottomrule
\end{tabular}
\end{table}

\subsection{Ablation Study}
To isolate the contribution of each component we evaluate three
variants of our model under identical training settings on
\texttt{shapes\_rotation} (Table~\ref{tab:ablation}). Variant~A uses
the event branch alone; variant~B adds the IMU and fuses the two
modalities by concatenation; variant~C replaces the concatenation with
a cross-attention mechanism. Adding the inertial modality
(A~$\rightarrow$~B) reduces the median position error from $0.0112$~m
to $0.0047$~m and the orientation error from $1.780^\circ$ to
$1.009^\circ$, i.e.\ relative improvements of $58\%$ and $43\%$
respectively, for only $23\%$ more parameters. This is the central
result of our study: the inertial signal, freely available on the
sensor and ignored by prior work, accounts for a substantial share of
the achievable accuracy. In contrast, replacing the concatenation with
a cross-attention fusion (B~$\rightarrow$~C) leaves both errors
essentially unchanged while doubling the parameter count from
$0.379$M to $0.807$M. We therefore retain the concatenation-based
variant~B as our final model: the gain stems from the inertial
information itself, not from the sophistication of the fusion
mechanism.

\subsection{Inference Time and Efficiency}
Beyond accuracy, \method{} is designed to be deployable on
resource-constrained platforms. With $0.379$M parameters, it runs in $3.21$~ms per
sample on a single CPU core, comfortably within the $5$~ms budget
implied by the $200$~Hz sampling rate, and thus
sustains real-time operation without any GPU. This stands in contrast
to heavier baselines such as SP-LSTM ($135$M parameters) and confirms
that the accuracy of \method{} does not come at the price of
efficiency.

\section{DISCUSSION AND LIMITATIONS}
\label{sec:discussion}

Our experiments show that the inertial modality yields a large and
consistent improvement under the random-split protocol. Interestingly,
the gain is comparable in position and in orientation, even though the
accelerometer most directly constrains roll and pitch through the
gravity vector. We attribute this to the geometry of the benchmark
sequences, in which rotational and translational motion are strongly
coupled, so that a better-constrained orientation also improves the
regressed position.

Our method has two main limitations. First, it regresses the absolute
pose without an explicit hierarchical encoding of the scene's spatial
structure. Under the more challenging \emph{novel split}, where the
test window is temporally disjoint from training, this leads to
noticeably weaker generalization than hierarchical models such as
PEPNet. Incorporating a hierarchical spatial encoder, while preserving
the lightweight inertial fusion, is a promising direction for future
work. Second, for computational reasons the two densest sequences were
evaluated with fewer windows than the others; a full-resolution
evaluation on dedicated hardware would further strengthen the results.

\section{CONCLUSION}
\label{sec:conclusion}

We revisited a design choice shared by all prior learning-based
event-camera relocalization methods discarding the on-board
IMU and showed that this freely available inertial modality
substantially improves 6-DoF pose estimation. Our lightweight
\method{} fuses a point-cloud event encoder with a compact inertial
branch and matches or surpasses the state-of-the-art on the DAVIS RPG
benchmark while using roughly half the parameters and running in real
time on a CPU. An ablation study confirms that the gain stems from the
inertial information itself rather than from an elaborate fusion
mechanism. We hope these results encourage the community to treat the
IMU as a first-class input for event-based relocalization.


\printbibliography

\end{document}